\documentclass[sigconf]{acmart}
\renewcommand\footnotetextcopyrightpermission[1]{} 
\usepackage{cite}
\usepackage{amsmath,amsfonts}
\usepackage{algorithm}
\usepackage{algorithmic}

\usepackage{caption}
\usepackage{subfigure}
\usepackage{graphicx}
\usepackage{textcomp}
\usepackage{amsmath}
\usepackage{cases}
\usepackage{xcolor}
\usepackage{verbatim}
\usepackage{multirow}
\usepackage{booktabs}
\usepackage{mathrsfs}

\AtBeginDocument{%
  \providecommand\BibTeX{{%
    \normalfont B\kern-0.5em{\scshape i\kern-0.25em b}\kern-0.8em\TeX}}}

\setcopyright{acmcopyright}
\copyrightyear{2018}
\acmYear{2018}
\acmDOI{10.1145/1122445.1122456}

\copyrightyear{2021}
\acmYear{2021}
\setcopyright{rightsretained}
\acmConference[MRC'21]{The 1st International Workshop on Machine Reasoning}{March 12, 2021}{Jerusalem, Israel}
\acmBooktitle{Proceedings of the 1st International Workshop on Machine Reasoning (MRC'21), March 12, 2021, Jerusalem, Israel}

\begin{document}

\title{Metapaths-guided Neighbors-aggregated Network for Heterogeneous Graph Reasoning}

\author{Bang Lin}
\affiliation{%
  \institution{Alibaba Group}
}
\email{linbang.lb@alibaba-inc.com}

\author{Xiuchong Wang}
\affiliation{
  \institution{Alibaba Group}
  }
\email{xiuchong.wxc@alibaba-inc.com}

\author{Yu Dong}
\affiliation{
  \institution{Alibaba Group}
  }
\email{dongyu.dy@alibaba-inc.com}

\author{Chengfu Huo}
\affiliation{
  \institution{Alibaba Group}
  }
\email{chengfu.huocf@alibaba-inc.com}

\author{Weijun Ren}
\affiliation{
  \institution{Alibaba Group}
  }
\email{afei@alibaba-inc.com}

\author{Chuanyu Xu}
\affiliation{
  \institution{Alibaba Group}
  }
\email{tracy.xcy@alibaba-inc.com}


\begin{abstract}
Most real-world datasets are inherently heterogeneous graphs, which involve a diversity of node and relation types. Heterogeneous graph embedding is to learn the structure and semantic information from the graph, and then embed it into the low-dimensional node representation. Existing methods usually capture the composite relation of a heterogeneous graph by defining metapath, which represent a semantic  of the graph. However, these methods either ignore node attributes, or discard the local and global information of the graph, or only consider one metapath. To address these limitations, we propose a \textit{Metapaths-guided Neighbors-aggregated Heterogeneous Graph Neural Network(MHN)} to improve performance. Specially, MHN employs node base embedding to encapsulate node attributes, BFS and DFS neighbors aggregation within a metapath to capture local and global information, and  metapaths aggregation to combine different semantics of the heterogeneous graph.  We conduct extensive experiments for the proposed MHN on three real-world heterogeneous graph datasets, including node classification, link prediction and online A/B test on Alibaba mobile application. Results demonstrate that MHN performs better than other state-of-the-art baselines.
\end{abstract}




\maketitle

\section{Introduction}

Real-world datasets usually exist in the form of graph structure, such as social networks[1], citation networks[2], knowledge graphs[3], especially recommendation systems[4] where nodes and edges represent objects and relationships, respectively. Taking an example, users and items in recommendation system can be represented as nodes while relationships such as purchases and clicks can be represented as edges, so that we can turn the recommendation dataset into a graph structure. Because graph is a high-dimensional non-Euclidean structure, it is difficult to model by traditional machine learning methods. Therefore, it is helpful to represent nodes by low-dimensional dense vectors, which can be the input of other machine learning models.

There have been many graph embedding methods, which are divided into two different solutions generally. For methods based on random walk, such as Deepwalk[5], Node2vec[6] and so on, sequences generated by random walk are fed into the skip-gram model to learn node embedding. However, with the rapid development of neural networks, methods based on graph neural network(GNN) have become more widely used, which learn the node embedding using specially designed neural layers. Methods like GCN[7], GAT[8], GraphSAGE[9] and other variants, they perform convolution operations on the graph or apply attention mechanism to generate more reasonable node representations.

Although the above methods have achieved state-of-the-art results in graph embedding learning, their input is homogeneous graph, which consist one edge type and one node type. Many real-world datasets are heterogeneous graphs, which consist of various types of nodes and edges. For example, an E-commerce recommendation graph consists at least two types of nodes, namely \textit{user} and \textit{item}. At the same time, different types of nodes have different attributes. Attributes of \textit{user} node may include age, sex and address while \textit{item} node attributes may consist of price, brand, category, and so on. Due to the heterogeneity of graphs, it is a challenge for GNNs to encoder the complex information into low-dimensional vectors.

Since metapath can extract relations between different types of nodes in heterogeneous graphs, most of heterogeneous graph embedding methods are based on metapath. Metapath is an ordered sequence of node and edge types, which represents a semantic space of the graph. For example, in E-commerce recommendation graph, metapath \textit{user-item-user} means the User-based collaborative filtering, while \textit{item-user-item} means the Item-based collaborative filtering. Also, metapath can guide the way to sample the heterogeneous graph and obtain neighbors.

Currently, there are many methods using metapath to generate node representation  of heterogeneous graph, but they still have some limitations. (1) Some methods do not make use of the attributes of nodes, resulting in the lack of rich information, such as Metapath2vec[10], HERec[11]. (2) Some methods do not consider local and global information of nodes, which are important to the generation of node embedding(e.g. HAN[12], GATNE[13]). (3) Although nodes in different semantics have different meanings, some methods adopt one metapath to embed the heterogeneous graph, ignoring the importance of multiple semantic spaces(e.g. Metapath2vec[10], EGES[14]).

In order to address these limitations, we propose a Metapaths-guided Neighbors-aggregated Heterogeneous Graph Neural Network(MHN) model for heterogeneous graph embedding learning. Through applying node base embedding by attributes transformation, aggregation within one metapath and aggregation among matapaths, MHN can address these limitations. Specifically, MNH first adopts type-specific linear transformations to project attributes of different types of nodes, aiming to transform them into the same latent vector space. Then, MHN applies metapath-guided neighbors aggregation for each metapath. During extracting local information from BFS neighbors and global information from DFS neighbors, MHN weighted sums them and obtains the representation of target node under the current semantic. In this way, MHN captures structural and semantical information of the heterogeneous graph. Finally, MHN conducts aggregation among metapaths using the attention mechanism, with the aim of fusing latent vectors obtained from multiple metapaths into the final node embedding. Therefore, MHN can learn the comprehensive semantics in the heterogeneous graph.

The contribution of this paper lies in three aspects:
\begin{itemize}
	\item We propose an end-to-end model MHN for heterogeneous graph embedding, which is a novel metapath aggregated graph neural network.
	
	\item MHN extracts local and global information under the guidance of a single metapath, and applies attention mechanism to fuse different semantic vectors. MHN supports both supervised and unsupervised learning.
	
	\item We conduct extensive experiments on the DBLP dataset for node classification task, as well as on the Amazon and Alibaba datasets for link prediction task to evaluate the performance of the proposed model. Moreover, we conduct online A/B test on Alibaba mobile application. Results show that representations generated by MHN performs better than other state-of-the-art methods consistently.
\end{itemize}

The rest of this paper is organized as follows. Section 2 introduces the related work. Section 3 describes some preliminary knowledge. Then we proposed the heterogeneous graph embedding method in Section 4. Experiments and analysis are shown in Section 5. Finally, we conclude the paper in Section 6.

\section{Related Work}
In this section, we will review the related studies about graph representation learning related to the proposed model. These methods are organized into four subsections: Homogeneous Graph Embedding methods, Homogeneous GNN methods, Heterogeneous Graph Embedding methods and Heterogeneous GNN methods.

\textbf{Homogeneous Graph Embedding methods}. The goal of these methods is to learn a low-dimensional representations for each node from homogeneous graph, which can be used for many downstream task directly.
DeepWalk[5] is a model for learning latent representation, which applies random walk to obtain node sequences and feeds them into skip-gram model to generate representations.
LINE[15] learns node representations on large-scale graph, which summaries local and global information through first-order and second-order proximities.
Node2vec[6] designs a biased random walk to explore diverse neighborhoods and maximize the likelihood of preserving network neighborhoods of nodes. 
SDNE[16] applies an autoencoder structure to optimize both the first-order and second-order similarities.

\textbf{Homogeneous GNN methods}. These methods are mainly built by homogeneous graph convolution and can be used for both supervised and unsupervised learning. GCN[7] generates the node embedding by graph convolution, which is performed in the graph Gourier domain.
GAT[8] introduces the attention mechanism into the graph convolution, and assigns different weights to neighboring nodes to update the node representation.
GraphSage[9] is a inductive learning method. By training the aggregation function, it can merge features of neighborhoods and generate the target node embedding.

\textbf{Heterogeneous Graph Embedding methods}. Unfortunately, most of above studies focus on the homogeneous graph and cannot be used directly in heterogeneous graph. Nowadays, there are more and more studies about heterogeneous graph embedding.
Metapath2vec[10] proposes to use metapath guided random walk to sample the heterogeneous graph and obtain several node sequences.
EGES[14] is proposed to solve above problem. Through attention mechanism, it can merge attribute information into node embedding, which have achieved good improvement in CTR prediction task.
HIN2vec[17] captures the rich semantics embedded in heterogeneous graph by predicting whether there is a metapath between nodes.

\textbf{Heterogeneous GNN methods}. Due to the complexity of heterogeneous graph, homogeneous GNN methods cannot be applied directly. There are many researches on how to introduce graph convolution into heterogeneous graph.
HAN[12] proposes a node-level attention layer to aggregate neighbors of target node features and a semantic-level attention layer to merge different semantic representations.
GATNE[13] solves the problem of embedding learning for the heterogeneous graph with attributes.
HERec[11] proposes an embedding method for heterogeneous graph and applies to the recommendation scene by matrix factorization.

However, heterogeneous graph embedding methods introduced above have the limitation of ignoring the local and global information. Although they have achieved some results in several datasets, we believe that there is still room for improvement by fully utilizing the information, which promotes us to study the optimal method of embedding for heterogeneous graph. 

\section{PRELIMINARIES}

In this section, we give definitions of important terms related to heterogeneous graph and show them in Figure 1. 

\textbf{Definition 3.1 Heterogeneous graph}[18]. A heterogeneous graph is denoted as a graph $G=(\mathcal{V},\epsilon)$, consisting of a node set $\mathcal{V}$ and a link set $\epsilon$, which is also associated with a node type mapping function $\varphi:\mathcal{V} \xrightarrow{} \mathcal{A}$ and a link type mapping function $\psi : \epsilon \xrightarrow{} \mathcal{R}$. $\mathcal{A}$ and $\mathcal{R}$ denote the sets of predefined node types and link types, where $|\mathcal{A}| + |\mathcal{R}| > 2$.

\textit{Example.} As shown in Figure 1(a), we construct a heterogeneous graph to model the E-commerce. It consists of two type of nodes(\textit{User}($U$) and \textit{Item}($I$)) and two relations, which are user click item relation(u-i) and the similarity of items relation(i-i).

\textbf{Definition 3.2 Metapath}[19]. A metapath $p$ is defined as a path in the form of $p=A_1\xrightarrow{R_1}A_2\xrightarrow{R_2}A_3 \cdots \xrightarrow{R_l}A_{l+1}$, which describes a composite relation $R=R_1 \circ R_2 \cdots R_l$ between objects $A_l$ and $A_{l+1}$, where $\circ$ denotes the composition operator on relations. Different metapaths represent different semantics[20].

\textbf{Definition 3.3 Metapath instance}[21]. Given a metapath $p$ of a heterogeneous graph, we can sample the graph under the guidance of $p$ and obtain several node sequences, which is defined as metapath instance.

\textit{Example.} As is shown in Figure 1(c), under the guidance of metapath $p = \{\textit{U-I-I-U}\}$, we can sample the graph and get two metapath instances $user1-item1-item2-user2$ and $user1-item3-item4-user3$.

\textbf{Definition 3.4 Metapath based BFS Neighbors}. Given a node $u$ and a metapath $p$ in a heterogeneous graph, we can sample the the metapath and obtain metapath instance for $u$, which is denoted as $P(u)$. Metapath based BFS Neighbors $N_u^p$ of node $u$ is defined as the set of nodes which connect to the node $u$ directly from $P(u)$. Note that the node’s BFS neighbors does not include itself.

\textit{Example.} Taking Figure 1(d) as an example, given the metapath $p = \{\textit{U-I-I-U}\}$, the metapath based BFS neighbors of $user1$ is $\{item1,item3\}$ because we can get two metapath instances. Obviously, metapath based BFS neighbors can exploit the local information of the graph because these neighbors connect to target node directly. 

\textbf{Definition 3.5 Metapath based DFS Neighbors}. Given a node $u$ and a metapath $p$ in a heterogeneous graph, after sampling the graph and get metapth instance $P(u)$, we can randomly choose a node sequence $t \in P(u)$. Metapath based DFS Neighbors $M_u^p$ is defined as the set of nodes that appears on node sequence $t$. Note that the node’s DFS neighbors dose not include the first two nodes.

\textit{Example.} Taking Figure 1(d) as an example, given a metapath $p = \{\textit{U-I-I-U}\}$, we can randomly choose a node sequence $\{user1-item1-item2-user2\}$ from metapath instances. According to the definition, we remove the first two nodes and obtain the metapath based DFS neighbors of $user1$ is $\{item2, user2\}$.  Compared with BFS neighbors, DFS neighbors focus on the global information.

\textbf{Definition 3.6 Heterogeneous Graph Embedding}. Given a heterogeneous graph $G=\{ \mathcal{V}, \epsilon\}$, heterogeneous graph embedding is the task to learn the $d$-dimensional node representations $z_u \in \mathbb{R}^d, \forall u \in \mathcal{V}$, which can capture the structural and semantic information of the heterogeneous graph, where $d \ll |\mathcal{V}|$.

\begin{figure*}
	\centering
	\includegraphics[width=.95\linewidth]{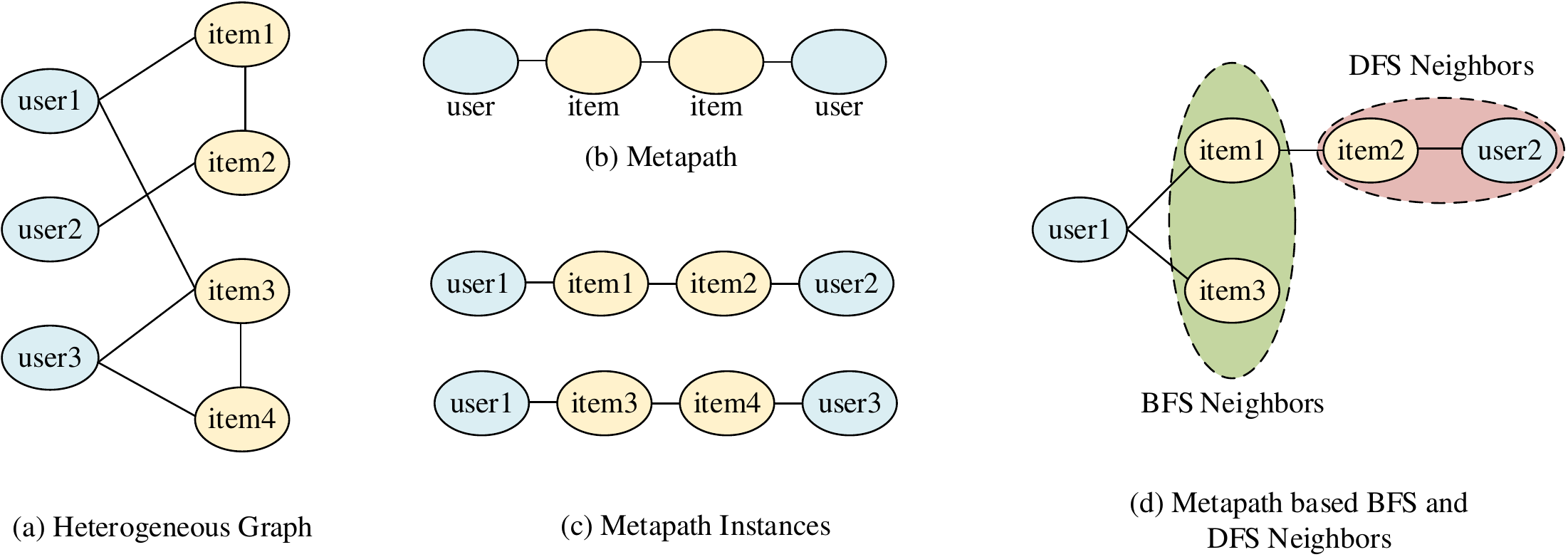}
	\caption*{ \textbf{Figure 1: Illustration of definitions. (a) A heterogeneous graph with two types of nodes(i.e., \textit{users, items}). (b) An example of metapath, User-Item-Item-User(\textit{UIIU}). (c) Metapath instances of the \textit{UIIU}. (d) The metapath based BFS and DFS neighbors of the \textit{UIIU}}} 
	\label{fg:state}
\end{figure*}

\section{THE PROPOSED MODEL}
In this section, we present a metapath guided Heterogeneous Graph Embedding Method, called MHN.

In order to make full use of node attributes and structure information of HIN, the proposed MHN model consists of three major components as is shown in Figure 2. Firstly, we propose a node embedding representation method combining node attributes. Then, we propose an attention based method to aggregate local and global information of HIN under a single semantic space. Finally, we propose a fusion model to merge the node embeddings under multiple semantics. We will present detailed illustration of the proposed model next.

\subsection{Node Base Embedding}
For large-scale networks, node id represents the node directly, which has a great impact on the node embedding. For example, Deepwalk[5] and Metapath2vec[10] feed node id into network to learn node embedding directly. Therefore, we apply a embedding lookup layer to get embedding from id. We have 

\begin{equation}
h_u^{id} = W_e \cdot u
\end{equation}

where $W_e \in \mathbb{R}^{|\mathcal{V}| \times d}$ is the parameter matrix, $u$ is the id of node, $h_u^{id}$ is the latent vector of the node. The role of this layer is to obtain the corresponding vector according to the node id. The parameter matrix $W_e$ is updated during the training process.

In real world graph, nodes are commonly attributed. Because  attributes represent the information of node, so it is important for heterogeneous graph. For example, the information of item contains characteristics such as brand, price, etc, which need to be reflected in node embedding. However, different kinds of node may have unequal feature dimension. Even for different types of nodes with the same feature dimension, their features have different meanings. So we can not simply use a matrix to transform attributes. Therefore, we need to design a method to map different types of node features into the same vector space.

In MHN, we  multiple transformation matrices to map different types of node attributes into the same space. For node $u \in \mathcal{V}_A$, we have
\begin{equation}
h_u^{att} = W_A \cdot x_u
\end{equation}

where $W_A$ is the parametric weight matrix for type A’s nodes, $x_u$ is the feature vector of node $u$, $h_u^{att}$ is the attribute transformed latent vector of node $u$.

Considering id and attributes, we can finally obtain the representation of the node by average these two vectors:
\begin{equation}
h_u = pooling(h_u^{id}, h_u^{att})
\end{equation} 

After applying these options, we can get the node latent vector containing id and attribute information in the same dimension. Then we will explore how to aggregate under the guidance of metapath. 

\subsection{Aggregation Within Metapath}
Given a single metapath $p_i \in \mathcal{P}$, the aggregation within metapath learns the local and global information through sampling the target node $u$ by BFS and DFS. Firstly, we sample the heterogeneous graph under the guidance of $p_i$ and get some paths started from $u$. Then, we let $N_u^{p_i}$ denote the BFS neighbors of $u$ under the metapath $p_i$. Through the function $f_\theta$, we can encode the neighbors and obtain the $h_{u,p_i}^{BFS}$.

\begin{equation}
h_{u,p_i}^{BFS} = f_\theta(h_v, v \in N_u^{p_i})
\end{equation}

where $f_\theta$ is the encoder function. We exam three functions:
\begin{itemize}
	\item \textbf{MEAN encoder} This function takes the mean of all the neighbors, thinking that all neighbors have the same contribution.
	\begin{equation}
	h_{u,p_i}^{BFS} = MEAN(h_v, v \in N_u^{p_i})
	\end{equation}
	
	\item \textbf{Weighted encoder} This function assigns different weights to neighboring nodes through $\beta$.
	\begin{equation}
	h_{u,p_i}^{BFS} = SUM( \beta \cdot [h_v, v \in N_u^{p_i}])
	\end{equation}
	
	\item \textbf{Non-linear encoder} The above two functions focus on the linear aggregation, which has limited expressive power in modeling complex relations. So we propose a non-linear function to enhance the representation of the relations through parameter matrix $W \in \mathbb{R}^{|h^{'} \times h^{'}}$, which is updated during the process of training.
	\begin{equation}
	h_{u,p_i}^{BFS} = \sigma ( W \cdot [h_v, v \in N_u^{p_i}])
	\end{equation}
	Where $\sigma()$ is non-linear function, i.e., sigmoid or relu. 
	
\end{itemize} 

Similarity, with the DFS neighbors of $u$ under the metapath $p_i$, we can encode $v \in M_u^{p_i}$ and obtain $h_{u,p_i}^{DFS}$.

After encoding the BFS and DFS information into vector representations, we adopt a simple attention mechanism to weighted sum two vectors related to target node $u$. The key idea is that BFS neighbors and DFS neighbors have different impacts on node representation. We can model this by learning a normalized importance weight $[\alpha_{1}, \alpha_{2}]$:
\begin{equation}
\begin{split}
e_1, e_2 &= h_u^T \cdot h_{u,p_i}^{BFS}, h_u^T \cdot h_{u,p_i}^{DFS} \\
\alpha_1, \alpha_2 &= \frac{ e^{e_1}} {e^{e_1} + e^{e_2}}, \frac{e^{e_2}} {e^{e_1} + e^{e_2}} \\
h_u^{p_i} = &\alpha_1 \cdot h_{u,p_i}^{BFS} + \alpha_2 \cdot h_{u,p_i}^{DFS}
\end{split}
\end{equation}

where $h_u$ is the representation of node $u$, $\alpha_1$ and $\alpha_2$ represents the relevance of target node between BFS information and DFS information.

In general, given the heterogeneous graph $G=(\mathcal{V}, \epsilon)$, node attributes $x_u, \forall u \in V$ and a set of metapaths $\mathcal{P}=\{p_1,…,p_{|\mathcal{P}|}\}$, aggregation within metapath of MHN generates $M$ metapath guided vectors for target node $u$, denoted as \{$h_u^{p_1},…,h_u^{p_|\mathcal{P}|}$\}. Each $h_u^{p_i}$ can be interpreted as the representation of $p_i$ metapath instance of node u, which reflects the semantic information of node $u$ under the metapah $p_i$.

\begin{figure*}
	\centering
	\includegraphics[width=.95\linewidth]{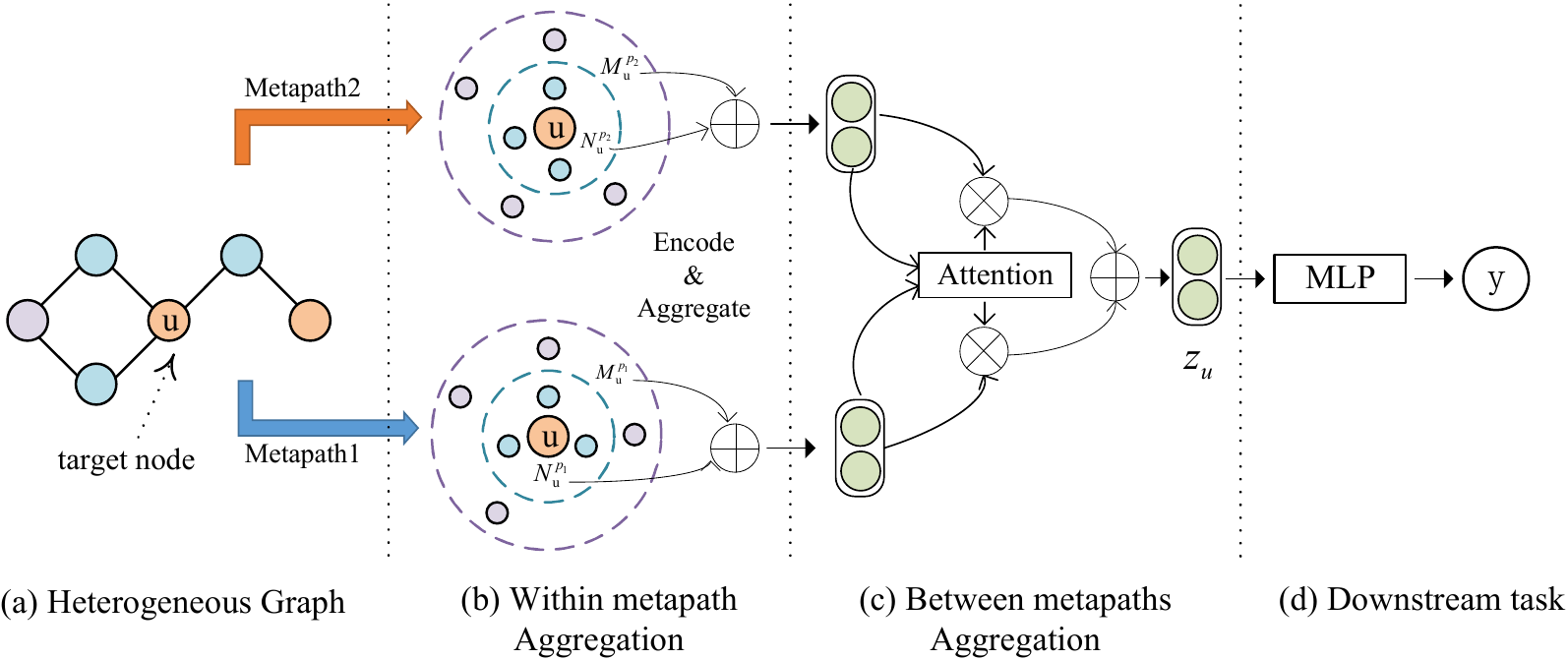}
	\caption*{ \textbf{Figure 2: The overall structure of proposed method MHN. (a) Sampling the heterogeneous graph and obtain metapath instances. (b) Encode BFS and DFS neigbors to summary local and global information, and aggregate them through weighted summation. (c) Calculate normalized attention weight for each metapath and generate embedding of node $u$. (d) Design the loss according to the downstream task and end-to-end optimize the model.}} 
	\label{fig:BLSTM}
\end{figure*}

\subsection{Aggregation Among Metapaths}

After aggregating DFS and BFS information to generate the final representation under a single metapath, we need to merge these different semantic information revealed by metapaths into a embedding vector. For $\forall u \in \mathcal{V}$, we have $|M|$ latent embeddings $\{h_u^{p_1},…,h_u^{p_M}\}$, where $M$ is the number of metapaths and $M=|\mathcal{P}|$. In order to obtain the final embedding, we assign different weights to different metapaths through the attention mechanism. The operation is reasonable because the optimization object may focus on different semantics.

We apply the attention mechanism to merge mebeddings of node $u$ under different semantics as follows:

\begin{equation}
\begin{split}
e_{p_i} &= q^T \cdot h_u^{p_i} \\
\beta_{p_i} &= \frac{e^{e_{p_i}}}{\sum_{p \in \mathcal{P}} e^p} \\
h_u &= \sum_{p_i \in \mathcal{P}} \beta_{p_i} \cdot h_u^{p_i}
\end{split}
\end{equation}

where $q$ is the parameterized attention vector which is updated in backpropogation, $\beta_{p_i}$ be interpreted as the importance of metapath $p_i$. 

Attention mechanism can also be extended to multi-heads self attention, which helps to stabilize the learning process and reduce the high variance. We first form all the embeddings of node $u$ into  matrix $H_u$ of shape $[M, d]$, where $M$ is the number of embeddings and $d$ is the embedding dimension. Then we calculate self-attention output under each head. Finally, we concatenate the output of each head as the embedding of node. Taking head 1 as an example, the calculating process is as follows:
\begin{equation}
\begin{split}
Q_1, K_1,V_1 &= H_u \cdot W^{Q_1}, H_u \cdot W^{K_1}, H_u \cdot W^{V_1}\\
h_{u,1} &= softmax(\frac{Q_1 \cdot K_1}{ \sqrt{d}}) \cdot V_1 \\
h_u &= concatenate_{k=1}^K h_{u,k}
\end{split}
\end{equation}

where $ W^{Q_0},  W^{K_0},  W^{V_0} \in \mathbb{R}^{d\times \frac{d}{K}}$, $K$ is the number of heads, $softmax = e^i / \sum_j e^j$, $d$ is the embedding size.

At last, we apply a fully connected layer to enhance the nonlinear fitting ability of the network and the output is the final embedding of node $u$:
\begin{equation}
z_u = \sigma(W \cdot h_u)
\end{equation}

\subsection{Metapaths Generation}
When using methods based on metapaths for heterogeneous graph embedding, we usually need to handcraft some metapaths which are adopted to sample the graph.  However, it is not trivial for human to find useful metapaths in a complex heterogeneous graph with multiple node or edge types. Therefore, we need to design an automatic generation method of metapath that does not rely on manual intervention, which can generate the most useful metapaths and sample as many nodes as possible.

MST[25] is proposed to select metapaths by using maximal spanning tree, which is instructive but ignore the applicability of metapaths. We propose a three stage approach to generate reasonable metapaths from heterogeneous graph automatically. In the first stage, we perform random walks on heterogeneous graph under a certain length and obtain lots of metapaths instances.  Secondly, through node type mapping and rule constraints, we can get hundreds of metapaths. According to unique demand, we can design scoring function and rank these metapaths to get the top K highest. Take Alibaba dataset as an example, we generate 4 million node sequences by setting sequence length to 10 and sampling 5 instances for each node. After filtering through rules, e.t. all node types must appear in the node sequence, we get almost 400 candidate metapaths. Due to the goal of training more nodes of video type, we design scoring formula in Eq. (12)
\begin{equation}
score(i) = \frac{log c(i)}{i.count(u) / i.count(v)} 
\end{equation}
where $c(i)$ represents the instances sampled by the i'th metapath, $i.count(u)$ and $i.count(v)$ means the number of \textit{user} and \textit{item} in the i'th metapath.

\subsection{Training}
After finishing the above components, we generate the embedding of each node, which can be applied in different downstream tasks. According to whether there are labels of data, we mainly divide the training process into two paradigms, supervised learning and unsupervised learning.

For supervised learning with node labels, we minimize the cross entropy loss and update the network parameters through backpropagation and gradient descent. The cross entropy loss of multi-classification for supervised learning is:
\begin{equation}
L=-\frac{1} {|\mathcal{V}|} \sum_{u \in \mathcal{V}} \sum_{c=1}^C y_{u,c} \cdot log p_{u,c} 
\end{equation}

where $\mathcal{V}$ is the set of training nodes, $C$ is the number of classes, $y_{u,c}$ is 1 if the label of $u$ is $c$ else 0, $p_{u,c}$ is the probability that $u^{'}$s label belongs to $c$ obtained by model.

For unsupervised learning without node labels, we can optimize the model by minimizing the following nce-loss function through negative sampling:
\begin{equation}
L=-\sum_{(u,v) \in S} log \sigma(z_u^T \cdot z_v) - \sum_{(u^{'},v^{'}) \in S^{-}}log \sigma (-z_{u^{'}}^T \cdot z_{v^{'}})
\end{equation}

where $\sigma(\cdot)$ is the sigmoid funchtion, $S$ is the positive node pairs, $S^{-}$ is the negative node pairs.

Through above model, we can not only aggregate the DFS and BFS information within a single metapath of the HIN, but also merge the different semantics represented by metapaths into the final embedding. The algorithm is shown in Algorithm1.

\begin{algorithm}[h]
	\caption{MHN forward propagation}
	\begin{flushleft}
		\hspace*{0.0in} {\bf Input:} The heterogeneous graph $G=(\mathcal{V}, \epsilon)$ \\
		\qquad \quad node features $\{x_u, \forall u \in \mathcal{V}\}$, \\
		\qquad \quad node types $\mathcal{A}=(A_1,…,A_{|\mathcal{A}|})$, \\
		\qquad \quad metapaths set $\mathcal{P}=\{p_1,…,p_{|\mathcal{P}|} \}$ \\
		\hspace*{0.0in} {\bf Output:} The node embeddings $\{z_u, \forall u \in \mathcal{V}\}$
	\end{flushleft}
	
	\begin{algorithmic}[1]
		\FOR{each node type $A \in \mathcal{A}$ } 
		\FOR{node $u \in \mathcal{V}_{A}$} 
		\STATE Get node id information $h_u^{id}$ and attributes transformation $h_u^{att}=W_A \cdot x_u$
		\STATE Calculate node representation $h_u = mean(h_u^{id}+h_u^{att})$
		\FOR{metapath $p \in \mathcal{P}_A$}
		\STATE Aggregate nodes in $N_u^p$, $M_u^p$ to obtain vectors $h_{u,p}^{BFS}$, $h_{u,p}^{DFS}$
		\STATE Calculate the weight $\alpha_1, \alpha_2$ for two vectors
		\STATE Obtain $h_u^p=\alpha_1 \cdot h_{u,p}^{BFS} + \alpha_2 \cdot h_{u,p}^{DFS}$ 
		\ENDFOR
		\STATE Calculate the weight $\beta_p$ for each metapath $p \in P_A$
		\STATE Merge the embeddings from all metapths:
		\STATE $h_u = \sum_{p \in \mathcal{P}_A} \beta_p \cdot h_u^p$
		\ENDFOR        
		\ENDFOR
		\STATE $z_u = \sigma(W_0 \cdot h_u), \forall u \in \mathcal{V}$
		\STATE return $z_u$
		
	\end{algorithmic}
\end{algorithm}

\section{experiments}
In this section, we present several experiments to demonstrate the effectiveness of the model we proposed in this paper. We verify the model on both offline and online datasets.

\subsection{Datasets}
In order to evaluate the performance of MHN as compared to state-of-the-art baselines, we adopt two widely used heterogeneous graph datasets and collect a real-world dataset from Alibaba mobile application from Android and IOS online. Specifically, the DBLP dataset is used in the experiments for node classification and visualization. Amazon and Alibaba datasets are used in the experiment for link prediction. The details of these datasets are shown in Table 1.
\begin{itemize}
	\item \textbf{DBLP} is a computer science bibliography website, which we  adopt a subset of DBLP extracted by [23]. The heterogeneous graph contains 4057 author nodes, 14328 paper nodes, 20 conference nodes, 19645 pa(paper to author) links and 14328 pc(paper to conference) links. These nodes are divided into four classes(Database, Data Mining, Artificial Intelligence and Information Retrieval). Paper's feature is made by its terms. Author's and publication's feature is described by a bag-of-words representation of their papers' terms. For supervised learning tasks, we divide author nodes into training, validation, test sets of 3245(80.00\%), 406(10.01\%), 406(10.01\%).
	
	\item \textbf{Amazon} includes product metadata and links between products. We adopt a subset of Amazon extracted by [24], in which we only use the product metadata of Electronics category. We build a heterogeneous graph including the co-viewing and co-purchasing links between products, and the product attributes include the price, sales-rank, brand and category with one-hot processing. For unsupervised learning tasks, we divide the dataset into training, validation, test sets of 3475(74.46\%), 398(8.53\%), 794(17.01\%).
	
	\item \textbf{Alibaba} consists of four types of links including user-click-item, user-click-video, similarity relation between items, parallelism relation between item and video with three node types \textit{user, item, video}, which is sampled from the log of Alibaba mobile application. We build a heterogeneous graph by sampling several active users and their behaviors with items and videos. Under the guidance of section 4.4, we generate three metapaths, who's sampled nodes can cover 96\% all nodes. For unsupervised learning tasks, we divide the dataset into training, validation, test sets of 5800(80.00\%), 725(10.00\%), 725(10.00\%).
\end{itemize}

\begin{table}[!htbp]
	\centering
	\caption{Datasets Statics}
	\begin{tabular}{@{}cccc@{}}
		\toprule
		Dataset & Node            & Edge & Metapath \\ \midrule
		DBLP &
		\begin{tabular}[c]{@{}c@{}}author(A):4057\\ paper(P):14328\\ conference(C):20\end{tabular} &
		\begin{tabular}[c]{@{}c@{}}P-A:19645\\ P-C:14328\end{tabular} &
		\begin{tabular}[c]{@{}c@{}}APA\\ APCPA\end{tabular} \\ \midrule
		Amazon  & Product(P):3475 &  \begin{tabular}[c]{@{}c@{}}P$\stackrel{viewing}{\longrightarrow}$P:2683\\ P$\stackrel{purchasing}{\longrightarrow}$P:791\end{tabular}    &   \begin{tabular}[c]{@{}c@{}}P$\stackrel{1}{\longrightarrow}$P\\ P$\stackrel{2}{\longrightarrow}$P\end{tabular}       \\ \midrule
		Alibaba &
		\begin{tabular}[c]{@{}c@{}}user(U):2785\\ item(I):2780\\ video(V):2716\end{tabular} &
		\begin{tabular}[c]{@{}c@{}}U-I:2935\\ I-V:1380\\ U-V:2935 \\ I-I:8569 \end{tabular} &
		\begin{tabular}[c]{@{}c@{}}UIU\\ UIVIU\\ UVU\\ IUI\\ IUVUI\\ IVI \\ IIUIUVUVUI \\ IVIVIVIUII \\ IUIIUVUVUI \end{tabular} \\ \bottomrule
	\end{tabular}
\end{table}

\subsection{Comparing Methods}
We categorize different graph embedding methods into four groups and compare MHN against these methods. The overall embedding size is set to 200.

\textbf{Homogeneous Graph Embedding Methods.} The compared methods include Deepwalk[5], LINE[15] and node2vec[6]. As these methods can only deal with Homogeneous graph, so we ignore the heterogeneity of graph and treat datasets as homogeneous.
\begin{itemize}
	\item \textbf{Deepwalk} is a approach for learning latent representations of vertices in a network, using random walks to learn embeddings by treating walks as the equivalent of sentences. 
	
	\item \textbf{LINE} optimizes a objective function that preserves both the local and global network structures using proposed edge-sampling algorithm. 
	
	\item \textbf{Node2vec} is an algorithm framework for learning continuous feature representations for nodes in homogeneous network. We unify the feature of different kinds of nodes into the same dimension. 
\end{itemize}

\begin{table*}[!htbp]
	\centering
	\caption{Performance comparison (\%) on DBLP dataset for node classification task}
	\begin{tabular}{|c|c|c|c|c|c|c|c|c|c|c|}
		\hline
		\multirow{2}*{Dataset} & \multirow{2}*{Metrics} & \multirow{2}*{Train(\%)} & \multicolumn{4}{|c|}{Unsupervised} & \multicolumn{4}{|c|}{Supervised}
		\\
		\cline{4-11} 
		~ & ~ & ~ & Deepwalk & Node2vec & LINE & Metapath2vec & GCN & GAT & HAN & MHN
		\\
		\hline 
		\multirow{8}{*}{DBLP} & \multirow{4}{*}{Micro-F1} & 20 & 84.35 & 89.71 & 88.61 & 89.49 & 89.76 & 90.31 & 91.23 & \textbf{92.16} \\
		\cline{3-11}
		~ & ~ & 40 & 86.35 & 89.85 & 89.12 & 90.31 & 90.43 & 90.91 & 91.75 & \textbf{92.48} \\
		\cline{3-11}
		~ & ~ & 60 & 86.49 & 90.13 & 89.68 & 90.53 & 90.82 & 91.05 & 92.01 & \textbf{92.94} \\
		\cline{3-11}
		~ & ~ & 80 & 86.86 & 90.88 & 89.75 & 91.01 & 90.83 & 91.18 & 92.37 & \textbf{93.29} \\
		\cline{2-11}
		~ & \multirow{4}{*}{Macro-F1} & 20 & 82.49 & 89.27 & 88.36 & 88.97 & 89.61 & 89.68 & 90.75 & \textbf{91.37} \\
		\cline{3-11}
		~ & ~ & 40 & 82.59 & 89.96 & 88.52 & 90.03 & 89.74 & 89.74 & 90.96 & \textbf{91.65} \\
		\cline{3-11}
		~ & ~ & 60 & 82.97 & 90.06 & 89.23 & 90.17 & 90.19 & 89.75 & 91.37 & \textbf{92.02} \\
		\cline{3-11}
		~ & ~ & 80 & 85.27 & 90.25 & 89.42 & 90.62 & 90.56 & 90.69 & 91.89 & \textbf{92.31} \\
		\hline
	\end{tabular}
\end{table*}

\textbf{Homogeneous GNNs.} These models focus on the structure of network and use graph convolution method to obtain information, including GCN[7] and GAT[8]. We test these models on metapath-based homogeneous graph and select the best result.
\begin{itemize}
	\item \textbf{GCN} obtains information through convolutional operations in Fourier domain for semi-supervised learning.
	
	\item \textbf{GAT} adopts attention mechanism to perform convolution in the homogeneous graph through masked self attention layer.
	
\end{itemize}

\textbf{Heterogeneous Graph Embedding Methods.} The compared methods include Metapath2vec[10]. These methods focus on the generation of node embedding in heterogeneous graph.
\begin{itemize}
	\item \textbf{Metapath2vec} adopts metapath-guided random walk to generate train sentences and feed them into skip-gram model and generate node embedding. We test multiple metapaths and select the best result. 
	
	
\end{itemize}

\textbf{Heterogeneous GNNs.} These methods include HAN[12] and GATNE[13]. Due to the the precise capture of heterogeneous information, they perform well in heterogeneous graph.
\begin{itemize}
	
	\item \textbf{HAN} uses attention mechanism to combine embeddings from different metapath-guided graph into one vector, capturing information from different emtapath. 
	
	\item \textbf{GATNE} focuses on the combination of different types of edges using attention mechanism. We test several GATNE variants and choose the best model.
\end{itemize}

For skip-gram based models, like Deepwalk, LINE, Node2vec, Metapath2vec, we set the window size to 5, walk length
to 10, walks per node to 20, and number of negative samples to
5. For GNNs, including GCN, GAT, HAN, GATNE, we apply Adam optimization with learning rate set to 0.01. These models are trained for 100 epochs and early stopping is set to 5. For models with attention mechanism, we set the number of attention head to 8 and attention dimension to 100.

\subsection{Node Classification}
We conduct experiment on the DBLP dataset to compare the performance of different methods on node classification task. We send node embeddings generated by different learning models into the Logistic Regression(LR) classifier with varying training proportions. In order to ensure fairness, all the data used for comparison comes from the test set, which both supervised learning and unsupervised methods have not trained. We compare the average \textit{Macro-F1} and \textit{Micro-F1} of different methods in Table 2.

As shown in Table 2, under different training proportions of DBLP dataset, MHN can achieve the best results over other learning methods. It is worth noting that, whether it is supervised or unsupervised learning, the method based on random walk performs better than the these based on GNN. This is because the DBLP dataset pays more attention to the connections between nodes, rather than the nodes themselves. Our method not only fully considers the global information, but also premeditates the local information, which ensures that the node embedding contains rich semantic information. The performance gain obtained by MHN over the best baseline(HAN) is about 0.42\%-0.93\% absolutely.

\subsection{Link Prediction}
Link prediction task is widely used to evaluate the quality of graph embeddings in both academia and industry. We also conduct experiments on the Amazon  and Alibaba datasets. We hide a set of edges as the test set and train on the remaining graph. For unsupervised learning models, we treat the connected links as positive node pairs and unconnected links as negative node pairs by minimizing the objective function described in Equation 14. Given the embedding $z_u$ and $z_v$, we calculate the probability that $u$ and $v$ are linked as following:
\begin{equation}
p_{u,v} = \sigma(z_u \cdot z_v)
\end{equation}

We use some commonly used metrics like \textit{the ROC curve}(ROC-AUC), \textit{the PR curve}(PR-AUC), \textit{the average precision}(AP) and \textit{the F1 score}.

\begin{table*}[!htbp]
	\centering
	\caption{Experiment results (\%) on Amazon and Alibaba datasets for link prediction task}
	\begin{tabular}{|c|c|c|c|c|c|c|c|c|}
		\hline
		\multirow{2}*{} & \multicolumn{4}{|c|}{Amazon} & \multicolumn{4}{|c|}{Alibaba}
		\\
		\cline{2-9} 
		~ & ROC-AUC & PR-AUC & F1 & AP & ROC-AUC & PR-AUC & F1 & AP
		\\
		\hline 			
		Deepwalk & 89.01 & 87.35 & 64.76 & 59.28 & 73.69 & 73.11 & 66.46 & 73.14 \\
		\hline 			
		Node2vec & 88.96 & 87.29 & 66.12 & 57.83 & 73.83 & 72.95 & 67.01 & 71.29 \\
		\hline 			
		LINE & 88.83 & 86.49 & 62.28 & 62.75 & 67.18 & 72.93 & 61.98 & 70.56 \\
		\hline
		Metapath2vec & 90.56 & 88.69 & 71.95 & 70.65 & 77.92 & 73.54 & 70.94 & 75.17 \\
		\hline 			
		GCN & 87.09 & 86.11 & 67.74 & 66.35 & 76.38 & 72.56 & 67.12 & 73.14 \\
		\hline 			
		GAT & 88.73 & 88.64 & 69.91 & 67.33 & 76.84 & 72.26 & 67.54 & 73.87 \\
		\hline
		HAN & 89.32 & 88.66 & 70.51 & 70.12 & 73.14 & 73.03 & 68.04 & 74.01 \\
		\hline
		GATNE & 89.27 & 88.04 & 69.79 & 70.06 & 74.11 & 72.89 & 68.23 & 73.91 \\
		\hline 			
		MNH & \textbf{92.02} & \textbf{89.81} & \textbf{73.15} & \textbf{71.46} & \textbf{79.37} & \textbf{74.92} & \textbf{71.85} & \textbf{75.38} \\
		\hline
	\end{tabular}
\end{table*}

From Table 3, we can see that MHN performs better than other comparison algorithms. The strongest traditional method here is Metapath2vec, which learns embedding from node sequences generated by random walk guided by one metapath. MHN achieves better scores than Metapath2vec, proving the importance of multiple semantics of heterogeneous graph. Based on the idea of multi-semantics fusion, MHN considers the influence of BFS and DSF neighbors of the target node, which helps achieve a relative improvement of around 8\% on Alibaba dataset over HAN. This result supports our claim that local and global information are critical to the node embeddings.

\subsection{Parameter Sensitivity}
We investigate the sensitivity of hyper-parameter in MHN, mainly the effect of embedding dimension. Figure 3 illustrates the performance of different methods when the embedding dimension changes. We can see that as the embedding dimension increases, the performance of models also increases, but the it drops when embedding dimension is either too small or too large. It can be conclude that the performance of MHN is relatively stable within the range of  embedding dimensions. Since the heterogeneous information cannot be identified, Deepwalk and GCN performs the worst. 

\begin{figure}
	\centering
	\includegraphics[width=.95\linewidth]{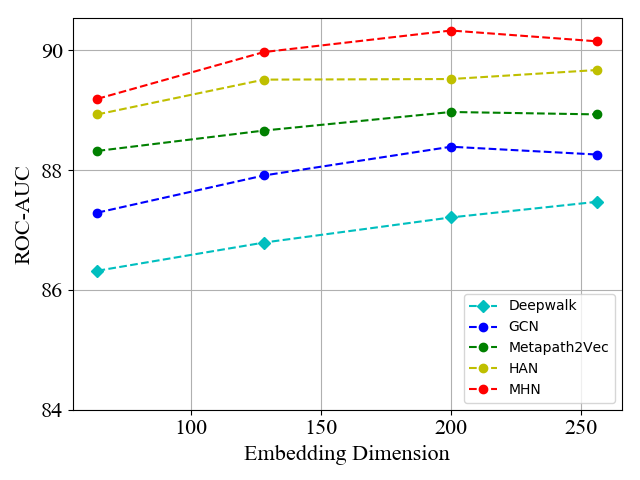}
	\caption*{ \textbf{Figure 3: The performance of different methods on Amazon dataset when changing  embedding dimensions.}} 
	\label{fg:BLSTM}
\end{figure}

\subsection{Visualization}
In addition to quantitative analysis of node embedding, we also adopt visualization method to qualitatively assess node embedding results. We randomly select four categories from DBLP dataset with 25 items under each category, and then project the embeddings of these nodes into a 2-dimensional space using t-SNE. We illustrate the visualization results of Deepwalk, Metapath2vec, HAN and MHN in Figure 4, where red and blue points indicate different category respectively.

Through visualization, we can intuitively tell the differences among learning ability of graph embedding methods for heterougeneous graph. As a traditional homogeneous graph representation learning method, Deepwalk cannot effectively divide these nodes into four groups. On the contrary, the heterogeneous model Metapah2vec can roughly distinguish these nodes. Because HAN fuse multiple semantics into node embedding, it achieves better performance. MHN method proposed in paper obtains the best embedding results, in which only a few nodes have errors and most of them are completely separated.

\begin{figure}
	\centering
	\subfigure[Deepwalk]{
		\centering
		\label{fig:subfig:a} 
		\includegraphics[width=0.48\linewidth]{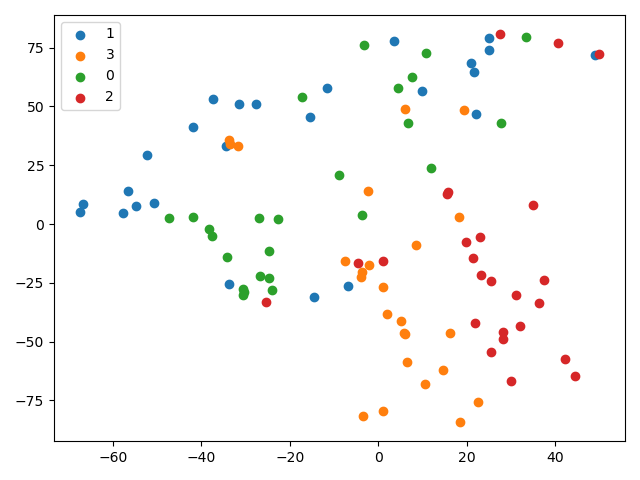}}
	\centering
	\subfigure[Metapath2vec]{
		\centering
		\label{fig:subfig:b} 
		\includegraphics[width=0.48\linewidth]{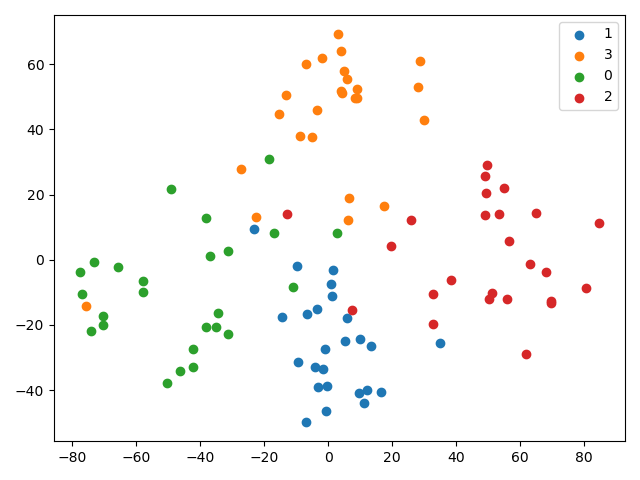}}
	\subfigure[HAN]{
		\centering
		\label{fig:subfig:c} 
		\includegraphics[width=0.48\linewidth]{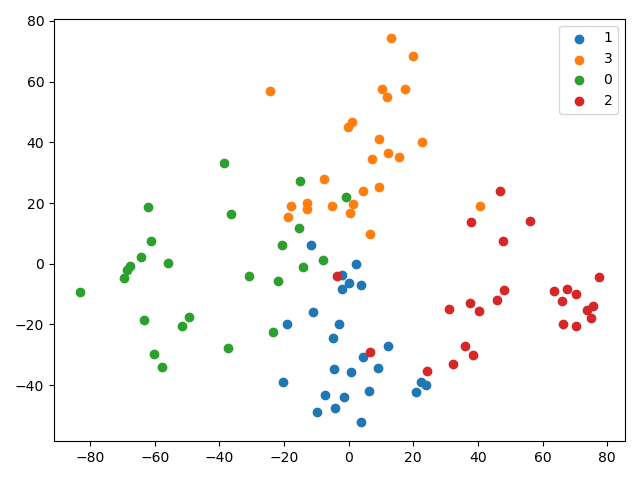}}
	\subfigure[MHN]{
		\centering
		\label{fig:subfig:d} 
		\includegraphics[width=0.48\linewidth]{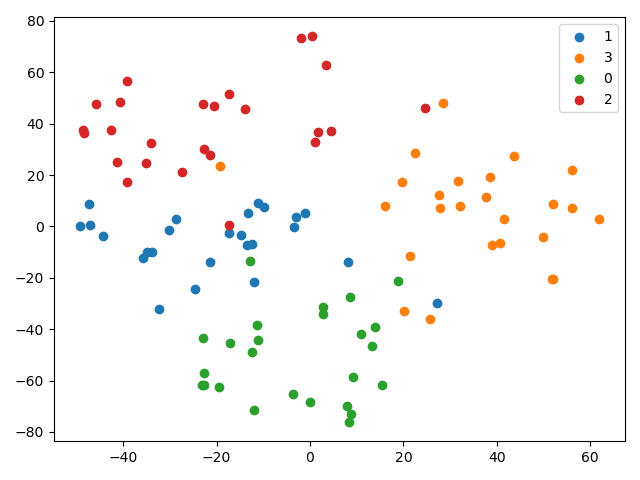}}    
	\caption*{\textbf{Figure 4: Embedding visualization of nodes in DBLP.} }
	\label{fig:BLSTM} 
\end{figure}

\subsection{Online A/B Test}
We deploy our inductive model MHN on Alibaba mobile application for it's  recall process of recommendation system. The training dataset has about half a million users and videos, a million items, with 3 million interactions among these nodes. We adopt MHN to generate embedding of each node under several metapaths. For each item, we apply K nearest neighbor (KNN) with Euclidean distance to obtain the top-50 videos that are most similar to the current item. Taking top-50 hit-rate as a goal, we compared the original method based on item-CF, Metapath2vec and MHN. The results demonstrate that MHN improves hit-rate by \textbf{2.93\%} and \textbf{6.71\%} compared to Metapath2vec and item-CF methods, respectively.

\section{Conclusion}
In this paper, we propose a metapaths-guided neighbors-aggregated Heterogeneous Graph Neural Network(MHN) method for heterogeneous graph node embedding learning, which can address three limitations mentioned above. MHN applies node base embedding  to transform node attributes and enrich node representation. In addition, aggregation within metapath can merge BFS and DFS neighbors to obtain local and global information of the target node, respectively. Finally, MHN adopts attention based algorithm in aggregation among metapaths to capture information in different semantics. Especially, we put forward several encode functions for neighbors aggregation and self-attention mechanism for vectors aggregation. In experiments, MHN achieves best results on three real-world datasets in node classification and link prediction task. Parameter sensitivity analysis illustrates the effect of embedding dimension. Visualization analysis shows the quality of node representation obtained by different methods directly. Online test in Alibaba mobile application proves the feasibility and effectiveness of MHN.

For the future work, we will consider about the dynamic heterogeneous graph node representation learning methods to adapt the changes in graph.


\end{document}